\documentstyle{nemlap}

\input{psfig}

\begin{document}

\title{New Methods, Current Trends and Software Infrastructure for NLP.}

\author{Hamish Cunningham, Yorick Wilks and Robert J. Gaizauskas}
\institute{Institute for Language, Speech and Hearing (ILASH), and \\
Department of Computer Science \\
University of Sheffield, UK}

\maketitle

% 
%	introduction.tex - GATE
%
%	Hamish Cunningham, 22/09/95
%
%	$Id: introduction.tex,v 1.3 1996/07/10 10:46:04 hamish Exp $

\begin{abstract}
The increasing use of `new methods' in NLP, which this conference series
exemplifies, occurs in the context of a wider shift in the nature and
concerns of the discipline. This paper begins with a short review of this
context and significant trends in the field. The review motivates and leads
to a set of requirements for support software of general utility for NLP
research and development workers. A freely-available system designed to meet
these requirements is described (called GATE - a General Architecture for
Text Engineering).
Information Extraction (IE), in the sense defined by the Message
Understanding Conferences (ARPA \cite{Arp95}), is an NLP application in
which many of the new methods have found a home (Hobbs \cite{Hob93}; Jacobs ed.
\cite{Jac92}).
An IE system based on GATE is also available for research purposes,
and this is described. Lastly we review related work.
\end{abstract}

\section{Introduction}
%\addcontentsline{toc}{section}{\protect\numberline{}{Introduction}}

The central theme of this paper is support software (or software {\em
infrastructure}) for NLP research and development (R\&D). This is not a new
concern -- witness for example the Alvey tools project (Grover et al.
\cite{Gro93}),
or the Core Language Engine (Alshawi ed. \cite{Als92}), or ALEP
(Simpkins \cite{Sim94}) -- although it is a subject that has
appeared in various forms in the literature of new methods in NLP (for
example, in this conference series: Nirenburg \cite{Nir94},
FGNLP-2; Cunningham et
al. \cite{Cun94},
NeMLaP-1). Recent trends in NLP (including, though not limitted
to, renewed interest in statistical methods, newly available corpora and
dictionary resources and tractable automatic learning algorithms) make
support software of particular current relevance. Recent work in the
European Linguistic Research and Engineering programme (in the MULTEXT project
(Thompson \cite{Tho95}; Ballim \cite{Bal95})) and the US TIPSTER
programme (in the TIPSTER architecture project (Grishman \cite{Gri95b}))
represent
responses to this situation. This paper reports work aimed at promoting and
synthesising the results of these programmes.

We begin by reviewing current trends in the field. This review motivates and
draws out a set of requirements for the provision of software infrastructure
for NLP R\&D. GATE (a General Architecture for Text Engineering) is a
freely-available system designed to meet these requirements, and is
described.
Information Extraction (IE), in the sense defined by the Message
Understanding Conferences (ARPA \cite{Arp95}), is an NLP application in
which many of the new methods have found a home (Hobbs \cite{Hob93}; Jacobs ed.
\cite{Jac92}).
An IE system based on GATE is also available for research purposes,
and this is described. Lastly we review related work.

% $Log: introduction.tex,v $
% Revision 1.3  1996/07/10  10:46:04  hamish
% final final version
%
% Revision 1.2  1996/07/09  16:55:03  hamish
% final submission
%
% Revision 1.1  1996/04/12  13:05:46  hamish
% as sent to Oflazer
%

% 
%	trends.tex - GATE
%
%	Hamish Cunningham, 22/09/95
%
%	$Id: trends.tex,v 1.2 1996/07/09 16:55:07 hamish Exp $

\section{\label{trends}Current trends in Language Engineering R\&D}

An increasing number of research and development
efforts have recently
positioned themselves under the banner {\em Language Engineering}
(LE). This signals a shift away from well-established labels such
as {\em Natural Language Processing} (NLP) and {\em Computational Linguistics}.
Examples include the renaming of UMIST's
{\em Department of Language and
Linguistics} (location of the {\em Centre for Computational Linguistics})
as the {\em Department of Language Engineering}, and the naming
of the European Commission's current relevant funding programme {\em
Language Engineering} (the previous programme was called {\em Linguistic
Research and Engineering}). The new journal of {\em Natural Language
Engineering} is another example%
\footnote{The editorial of the first issue also discusses the new name
(Boguraev, Garigliano, Tait \cite{Bog95}).}.

We shall argue here that this shift is more than simple TLA%
\footnote{TLA: three-letter acronym}%
-fatigue.
The new name reflects a change of emphasis within the field towards:

\begin{itemize}
\item increasing use of quantitative evaluation as a metric of research
achievement;
\item renewed interest in statistical language models and
automatically-generated resources;
\item increasing availability and use of large-scale resources (e.g.
corpora, machine-readable dictionaries);
\item a re-orientation of language processing research to large-scale
applications, with a comcomitant emphasis on predictability and conformance
to requirements specifications (i.e. emphasis on engineering issues).
\end{itemize}
Several commentators have characterised the
broad trend of AI approaches to language as tending towards the ``toy problem
syndrome'', expressing the view
that AI has too
often chosen to investigate artificial, small-scale
applications of the technology under development. 
%\resetparskip

For example,
one of the present authors began a large-scale Prolog grammar project in
1985 (Farwell, Wilks \cite{Far89}):
by 1987 it was perhaps the largest DCG (Definite Clause Grammar)
grammar anywhere, designed to cover a linguistically well-motivated test set
of sentences in English. Interpreted by a standard parser it was able to
parse completely and uniquely
virtually no sentence chosen randomly from a newspaper. We
suspect most large grammars of that type and era did no better, though
reports are seldom written making this point.

The mystery for linguists is how that can be: the grammar appeared to
inspection to be virtually complete -- it {\em had} to cover English, if
thirty years of linguistic intuition and methodology had any value. It is a
measure of the total lack of evaluation of parsing projects up to that time
that such conflicts of result and intuition were possible, a situation
virtually unchanged since Kuno's large-scale Harvard parser of the
1960's (Kuno, Oettinger \cite{Kun62})
whose similar failure to produce a single,
preferred, spanning parse gave rise to the AI semantics and knowledge-based
movement. The situation was effectively unchanged in 1985 but the response
this time around has been quite different, characterised by:
\begin{itemize}
\item
use of empirical methods with strict evaluation criteria;
\item
renewed interest in performance-based models of language, and a corresponding
renewal and extension of statistical techniques in the area;
\item
increased provision and reuse of large-scale data resources;
\item
greater emphasis on the
development of prototype applications of NLP technology to large-scale
problems.
\end{itemize}
With hindsight it may seem obvious that {\em computational} linguistics, in
the sense of computer programs that seek to exploit the results of linguistic
research to make computers do useful things with human language%
\footnote{There is, of course, at least one 
other sense, that of using computational
tools to aid linguistic research.},
should be subject to empirical criteria of effectiveness. The big problem,
of course, is determining precisely what the criteria of success should be.
Should we collect video tapes of Star Trek and measure our efforts in
comparison to the Enterprise's lucid conversational computer? There is now a
substantial literature on this question (Crouch, Gaizauskas, Netter
\cite{Cro95}),
and more practical solutions to the
evaluation problem have emerged in a number of areas.
%\resetparskip

Participants in the TIPSTER programme and
the MUC (Message Understanding Conference, an information extraction
competition)
and TREC
(Text Retrieval Conference, an information retrieval (or `document
detection') competition)
competitions (ARPA \cite{Arp93b}), for example,
build systems to do precisely-defined tasks on selected bodies of news
articles. Human analysts are employed to produce correct answers for some
set of previously unseen texts, and the systems run to produce machine
output for those texts. The performance of the systems relative to
human annotators is then measurable quantitatively.
Quantitative evaluation metrics bring numerically well-defined
concepts like precision and
recall, long used to evaluate information retrieval systems, to language
engineering%
\footnote{Machine Translation systems had been subject to
evaluation from its earliest days (ALPAC \cite{Alp66}),
%Lehrberger, Bourbeau 1988),
but this tradition did not spread further until recently.}%
.

A related phenomenon is the increasing use of statistical techniques in the
field (Jelinek \cite{Jel85}; Church \cite{Chu88};
Church, Mercer \cite{Chu93}).
Instead of an introspective
process of investigation into the underlying mechanisms by which people
process language (or, in Chomsky's terms, their {\em competence}),
statistical NLP attempts to build models of language as it exists in
practical use -- the {\em performance} of language.

Statistical methods have had significant successes, and the debate once
thought closed by Chomsky's `I saw a fragile whale' is now as open as it
ever has been. Most part-of-speech taggers now rely on statistics
and it seems possible that
parsers may also go this way,
though more conventional methods are also increasing in quality and
robustness.

It is possible that there is a natural ceiling to the advance of performance
models (Wilks \cite{Wil94}), but the point of relevance
for this paper is that the jury is still out on performance vs. competence.
Thus, as well as a host of competing linguistic and lexicographic theories,
LE is home to a thoroughgoing
paradigm conflict. Two important consequences ensue.

First, empirical measurement of the relative
efficacy of competing techniques is even more important.
Secondly, hybrid models are becoming common, implying a growing
need for the flexible combination of different techniques in single
systems. Numbers of techniques that have poor performance alone may
sometimes be combined to produce a whole greater than the sum of the parts
(Wilks, Guthrie, Guthrie, Cowie \cite{Wil92}).

In common with other software systems, LE components deploy both data and
process elements. The quality, quantity and availability of shared data
resources has increased dramatically during the late 1980s and 1990s.
(Extensive discussion of the repositories (LDC, CLR, MLSR etc.) of
corpora and lexicon resources and their holidings up to 1994 can be found in
(Wilks et al. \cite{Wil96}).
More recent developments concerning ELRA (the European
Linguistic Resources Association can be found in Elsnews 4.5 (November
1995).)

The sharing of processing (or algorithmic)
resources remains more limited (Cunningham, Freeman, Black \cite{Cun94}),
one key reason being that the integration and reuse of different components
can be a major task. 
Section \ref{trends} noted the increase in scale of the problems that
LE research
systems aim to tackle. In parallel with this trend, the overhead involved in
creating a full-scale IE system, for example, is also increasing. For many
research groups the costs are prohibitive. Any method for alleviating the
problems of reuse would make a signifcant contribution to LE research and
development.

On a smaller scale, the typical life-cycle of doctoral research in AI/NLP
is:
have an idea;
reinvent the wheel, fire and kitchen sinks to provide a framework for
the idea;
program the idea;
publish;
throw the system to the dogs / tape archivist / shelfware catalogue.
A framework which enabled relatively easy reuse of past work could
significantly increase research productivity in these cases.

With the increasing scale of LE systems, software engineering issues become
more important.
Just as the construction of the Golden Gate Bridge was a rather different
order of problem from that of laying a couple of planks across a farmland
ditch, the development of software capable of processing megabytes of text,
written by idiosyncratic wetware%
\footnote{Journalists.},
in short periods of time to
measurable levels of accuracy is a quite different game from that, say,
of providing natural
language interaction for the control of a robot arm that moves
blocks on a table top (Winograd \cite{Win72}).
The nuts and bolts are a lot bigger, and may even be of a
completely different fabric altogether.
This type of issue has been solved successfully in other areas of computer
science, e.g. databases. Failure to address software-level
robustness (as opposed to the robustness of the underlying NLP technology),
quality and efficiency will be a barrier to transferring LE technology from
the lab to marketplace.

Some other requirements relating to the technological foundations 
of these systems also arise.
Module interchangeability (at both the data and process levels), a kind of
`software lego' or `plug-and-play', would allow users to buy into LE
technology without tying them to one supplier. 
Also desirable are easy upgrade routes as technology
improves.
In addition to the reasons noted above, precise quantification of performance 
measures are also needed to foster
confidence in the capability of LE applications to deliver, and robustness
and efficiency for large text volumes are prerequisites for many
applications. 
Software multilinguality and operating system independence are also issues.
Finally, maximising cross-domain portability will favourably
impact delivery costs.

Our discussion of trends in LE concludes with 
two major LE application areas, Information Extraction (IE) and Machine
Translation (MT), which both exhibit the trends discussed above.
Recent years have seen significant improvements in the quality and
robustness of LE technology.
Rapid improvement in robustness (the ability to deal with any input) and
quality are evident in the leading systems.
In last year's MUC-6 competition
initial results indicate that named-entity recognition can now
be performed by machines to performance levels equal those of
people (ARPA \cite{Arp95}).
The result is that
applications of the technology to large-scale problems are increasingly
viable.

IE is intended to deal with the
rapidly growing problem of extracting meaningful information from
the vast amount of electronic textual data that threatens to engulf us.
Scientific journal abstracts, financial newswires, patents and patent
abstracts, corporate and government technical documentation, electronic mail
and electronic bulletin boards all contain a wealth of information of vital
economic, social, scientific and technical importance. The problem is that the
sheer volume of these sources is increasingly preventing the timely finding
of relevant information, a state of affairs exacerbated by the
explosive growth of the Internet.
Existing information retrieval (Salton \cite{Sal89})
solutions to this problem  are
a step in the right direction, and the industry supplying IR applications
can expect to continue in its current healthy state.
%
% IR, however, generally treats
% texts as unordered bags of words%
% \footnote{Exceptions here include ordered
% phrasal searches (e.g. the search term
% `information extraction' is different from `extraction information')
% and ordered proximity searches (e.g. `A within 10 words of B and before it',
% but are additions to the basic model, and stop short of any structural
% analysis of the texts.}%
% , and attempts no analysis of the meaningful content of texts.
%
IR systems, however, attempt no analysis of the meaningful content of
texts.
This is a
strength of the approach, leading to robustness and speed, but also a
weakness, as the information represented by the texts is retrieved in the
format of the texts themselves -- i.e. in the ambiguous and verbose medium
of natural language.
Extraction of information in definite formats is an obvious solution and one
which can only be achieved through the application of LE technology.

The IE community have been leaders in quantitative evaluation.
Statistical methods are widely used, but so is more conventional CL.
The need for systematic reuse
of both data and processing resources has been recognised, and work funded
to facilitate this, and the importance of software engineering matters noted.
A similar situation is evident in MT research: Machine Translation Vol.
8 nos. 1-2, Special Issue on Evaluation.

% $Log: trends.tex,v $
% Revision 1.2  1996/07/09  16:55:07  hamish
% final submission
%
% Revision 1.1  1996/04/12  13:05:55  hamish
% as sent to Oflazer

% 
%	gate.tex - GATE
%
%	Hamish Cunningham, 22/09/95
%
%	$Id: gate.tex,v 1.2 1996/07/09 16:55:01 hamish Exp $

\section{\label{gate}GATE}

GATE -- a General Architecture for Text Engineering -- is a project (funded
by the EPSRC under grant GR/K25267) that aims to address the infrastructural
needs of language engineering in the context of the trends described in the
previous section.

GATE is an {\em architecture} in the sense of providing a common
infrastructure for building LE systems. 
GATE is a {\em development environment} because it provides a variety of
data visualisation, debugging and evaluation tools (with point-and-click
interface), and a set of standardised interfaces to reusable components.

\begin{figure}[!htbp]
  \centerline{\psfig{figure=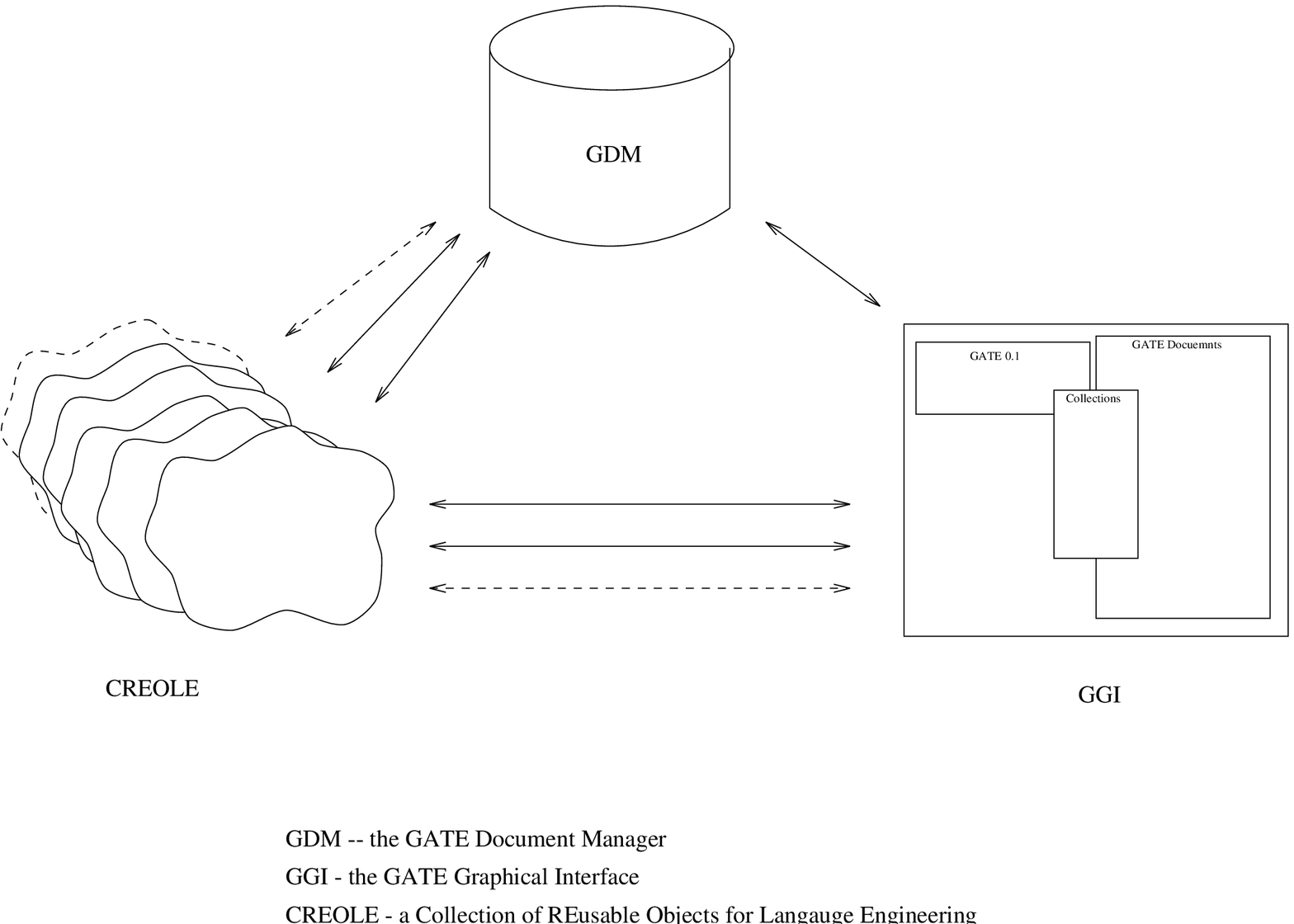,height=8cm}}
  \caption{\label{arch1}The three elements of GATE}
\end{figure}
GATE comprises three principal elements (figure \ref{arch1}):
\begin{itemize}
\item
a database for storing information about texts and a database schema
based on an object-oriented model of information about texts (the GATE
Document Manager -- GDM);
\item
a graphical interface for launching processing tools on data and viewing and
evaluating the results (the GATE Graphical Interface -- GGI);
\item
a collection of wrappers for algorithmic and data resources that
interoperate with the database and interface and constitute a Collection of
REusable Objects for Language Engineering -- CREOLE.
\end{itemize}
%\resetparskip

GDM is based on the TIPSTER document manager (Grishman \cite{Gri96}).
TIPSTER have
defined a neutral model of information associated with text (see below).
It is planned to enhance the SGML
capabilities of this model, possibly
by exploiting the results of the MULTEXT project
(we thank colleagues from ISSCO and Edinburgh for making available
documentation and advice on this subject).
See section \ref{related_work} for details of the relationship between GATE
and these projects.

GDM provides a central repository or server that stores all information an
LE system generates about the texts it processes. All communication between
the components of an LE system goes through GDM, insulating parts from each
other and providing a uniform API (applications programmer interface)
for manipulating the data produced by the
system.%
\footnote{Where very large data sets need passing between modules other
external databases can be employed if necessary.}\
Benefits of this approach include the ability to exploit the maturity and
efficiency of database technology, easy modelling of blackboard-type
distributed control regimes
and reduced interdependence of components.

GGI is in development at Sheffield. It is a graphical launchpad for LE
subsystems, and provides various facilities for viewing and testing results
and playing software lego with LE components: interactively stringing
objects into different system configurations.

All the real work of analysing texts (and maybe producing summaries of them,
or translations, or SQL statements\ldots) in a GATE-based LE system is done
by CREOLE modules.
Typically, a CREOLE object will be a wrapper around a pre-existing LE module
or database -- a tagger or parser, a lexicon or ngram index, for example.
Alternatively objects may be developed from scratch for the architecture --
in either case the object provides a standardised API 
to the underlying resources which allows access via GGI and
I/O via GDM. The CREOLE APIs may also be used for programming new objects.

The initial release of GATE will be delivered with a CREOLE set comprising a
complete MUC-compatible IE system (to begin with,
more of a pidgin than a creole!). Some of the objects will be based on
freely available software (e.g. the Brill tagger (Brill \cite{Bri94})),
while  others are derived from Sheffield's MUC-6 entrant, LaSIE%
\footnote{Large-Scale IE.}
(Gaizauskas, Humphreys, Wakao, Cunningham, Wilks \cite{Gai95b}).
This set is called VIE -- a Vanilla IE system. See
section \ref{vie} for an overview.

The recent MUC competition, the sixth, defined four IE tasks to be carried out
on Wall Street Journal articles. Sheffield's system did well, scoring in the
middle of the pack in general
and doing as well as the best systems in some areas.
Developing this system took 24 person-months, one significant element of which
was coping with the strict MUC output specifications. 
In GATE and VIE we
hope to provide an environment where groups can mix and match elements of
MUC technology from other sites (including ours)
with components of their own, thus allowing the benefits
of large-scale systems without the overheads. A parser developer, for
example, can replace the parser supplied with VIE.

Working with GATE/VIE, the researcher will from the outset reuse existing
components, the overhead for doing so being much lower than is
conventionally the case -- instead of learning new tricks for each module
reused, the common APIs of GDM and CREOLE mean only one integration
mechanism must be learnt. And as CREOLE expands, more and more modules and
databases will be available at low cost. 

As we built our MUC system it was often the case that we were unsure of the
implications for system performance of using tagger X instead of tagger Y,
or gazeteer A instead of pattern matcher B. In GATE, substitution of
components is a point-and-click operation in the GGI interface.
This facility supports hybrid systems, ease of
upgrading and open systems-style module interchangeability.
Of course, GATE does not solve all the problems involved in plugging diverse
LE modules together. There are two barriers to such integration:
\begin{itemize}
\item
incompatability of {\em representation} of information about text and the
mechanisms for storage, retrieval and inter-module communication of that
information;
\item
incompatability of {\em type} of information used and produced by different
modules.
\end{itemize}
GATE enforces a separation between these two and provides a solution to the
former based on the work of the TIPSTER architecture group.
Because GATE places no constraints on the linguistic formalisms or
information content used by CREOLE objects, the latter problem must be solved 
by dedicated translation functions -- e.g. tagset-to-tagset mapping -- and, in
some cases, by extra processing -- e.g. adding a semantic processor to
complement a bracketing parser in order to produce logical form to drive a
discourse interpreter. As more of this work is done we can expect the
overhead involved to fall, as all results will be available as CREOLE
objects. In the early stages Sheffield will provide some 
resources for this work in order to get the ball rolling, i.e. we will
provide help with CREOLEising existing systems and with developing interface
routines where practical and necessary. We are confident that integration
{\em is} possible (partly because we believe that differences between
representation formalisms tend to be exaggerated) -- and others share this
view, e.g. the MICROKOSMOS project (Beale, Nirenburg, Mahesh \cite{Bea95}),
which
seeks to integrate many types of knowledge source in a useable whole, as
well as the LexiCadCam experience at New Mexico (Wilks, Guthrie,
Slator \cite{Wil96}) which
sought to provide core lexical information as needed in a range of
user-specified formats.
%\resetparskip

GATE is also intended to benefit the LE system developer (which may be the
LE researcher with a different hat on, or industrialists implementing
systems for sale or for their own text processing needs).
A delivered system comprises a set of CREOLE objects, the GATE runtime
engine (GDM and associated APIs) and a custom-built interface (maybe just
character streams, maybe a Visual Basic Windows GUI, \ldots). The interface
might reuse code from GGI, or might be developed from scratch.

The LE user may upgrade by swapping parts of the CREOLE
set if better technology becomes available elsewhere. This model for the
commercialisation of LE technology is already begininning to operate in the
US, where a number of organisations are preparing TIPSTER-compatible modules
for sale or distribution for research.
(These organisations include NMSU, SRA, HNC,
University of Massachusetts, Paracell, Logicon (Dunning 1995, personal
communication).) All TIPSTER-compatibile modules will also work with GATE as
GATE itself is desinged to be a TIPSTER-compatible system. Thus the pool of
easily reusable LE resources available to researchers and developers using
GATE has the potential to become a large, rich set of modules from a good
proportion of the LE community world-wide. Also, it may well become the case
that organisations purchasing LE software will require TIPSTER compatability
(this will be true of US government organisations, for example).

GATE cannot eliminate the overheads involved with porting LE systems to
different domains (e.g. from financial news to medical reports). Tuning LE
system resources to new domains is a current research issue
(see also: the LRE ECRAN project).
The modularity of
GATE-based systems should, however, contribute to cutting the engineering
overhead involved.

%\subsection*{Collaboration using GATE and VIE}
%
%Sheffield will support collaborative work using GATE/VIE for LE research
%groups (typically academic groups), businesses with IE needs and
%producers of lexicons and dictionaries.  The three groups are
%{\em technology}, {\em data} and {\em resource providers} respectively,
%contributing CREOLE modules, test data (e.g. manually extracted information
%and the relevant source texts) and machine-readable language resources (e.g.
%dictionaries).
%The projected benefits for participants include:
%\begin{itemize}
%\item
%comparative quantitative evaluation of candidate technologies for IE;
%\item
%technology providers
%may specialise on components of the IE task, avoiding the overhead of
%providing a complete IE system while still working within the framework of
%a complete NLP application;
%\item
%data providers (typically industrial concerns) get access to  IE technology
%applied to their particular textual problem domains;
%\item
%resource providers can assess the performance of their products and increase
%the market for them by encouraging their use in applied LE systems.
%\end{itemize}
%Note that there will be no requirement to supply source code for contributed
%modules, and that intellectual property and other rights will be safeguarded
%by appropriate legal agreements.
%%\resetparskip

% $Log: gate.tex,v $
% Revision 1.2  1996/07/09  16:55:01  hamish
% final submission
%
% Revision 1.1  1996/04/12  13:05:44  hamish
% as sent to Oflazer

% 
%	vie.tex - GATE
%
%	Hamish Cunningham, 22/09/95
%
%	$Id: vie.tex,v 1.2 1996/07/09 16:55:09 hamish Exp $

\section{\label{vie}VIE, a Vanilla Information Extraction system}

GATE will be distributed
with a set of CREOLE objects that together implement a complete information
extraction system capable of producing results compatible with the MUC-6
task definitions. This CREOLE set is called VIE, a Vanilla IE
system, and it is intended that participating sites use VIE as the basis for
specialising on sub-tasks in IE. By replacing a particular VIE module -- the
parser, for example -- a participating group will immediately be able
to evaluate their specialist technology's potential contribution to
full-scale IE applications. Sheffield has access to the MUC-6 scoring tools
(and the PARSEVAL software) and will run periodic evaluations of various
VIE-based configurations.

The most recent MUC competition, MUC-6, defined four tasks to be carried
out on Wall Street Journal articles:
named entity (NE) recognition, the recognition and classification of definite
entities such as names, dates, places;
coreference (CO) resolution, the identification of identity relations
between entities (including anaphoric references to them);
template element (TE) construction, a fixed-format, database-like
enumeration of organisations and persons;
scenario template (ST) construction, the detection of specific relations
holding between template elements
relevant to a particular information need (in this case personnel joining
and leaving companies) and construction of a fixed-format structure
recording the entities and details of the relation.
VIE is an integrated system that builds up a single, rich model of a text
which is then used to produce outputs for all four of the MUC-6 tasks. Of
course this model may also be used for other purposes aside from MUC-6
results
generation, for example we currently generate natural language summaries of
the MUC-6 scenario results.

\section{\label{related_work}Related work}

\subsubsection*{MULTEXT}

MULTEXT (Ballim \cite{Bal95}; Thompson \cite{Tho95})
was an EC project, whose
objective was to produce tools for multilingual corpus
annotation and sample corpora marked-up according to the same standards used
to drive the tool development. Annotation tools were to perform
text segmentation, 
POS tagging, morphological analysis and parallel text alignment.
The project defined an architecture centred on a
model of the data passed between the various phases of processing implemented
by the tools.
Organisational problems have,
unfortunately, led to an early termination of the project, but the tools and
the architecture they run in should still be completed and distributed for
research purposes.

The MULTEXT architecture is based on a commitment to TEI-style
(the Text Encoding Initiative)
SGML encoding of information about text. The TEI defines standard tag sets
for a range of purposes including many relevant to LE systems.
Tools in a MULTEXT system communicate
via interfaces specified as SGML document type definitions (DTDs --
essentially tag set descriptions),
using character streams on pipes.
A tool selects what information it requires from its input SGML
stream and adds information as new SGML markup. An advantage here is a
degree of data-structure independence: so long as the necessary
information is present in its input, a tool can ignore changes to other
markup that inhabits the same stream -- unknown SGML is simply
passed through unchanged. A disadvantage is that although graph-structured
data may be expressed in SGML, doing so is complex (either via concurrent
markup, the specification of multiple legal markup trees in the DTD, or by
rather ugly nesting tricks to cope with overlapping, aka ``milestone tags'').
Graph-structured information might be present in the output of a parser, for
example, representing competing analyses of areas of text.

Another feature of MULTEXT is a set of abstract data types (ADTs)
for all tool I/O supported by a single shared API
(Application Program(mers') Interface)
for access to the types. An executive (the {\em tool shell}) glues tools
together in particular configurations according to user specifiactions. The
shell may extract sub-trees from SGML documents to reduce the I/O load where
tools only require a subset of a marked-up document.
The ADT set forms an object-oriented model%
\footnote{OO in the sense of using inheritance and data encapsulation.}
of the data present in a marked-up document. Example classes include {\bf
Sentence}, {\bf SentenceBlock} (sequence of sentences), {\bf LexicalWord}
(word plus definition from a lexicon). The ADT model reflects the type of
processing available in the tool set --
there is a type {\bf TaggedSentence}, for
example, but not a {\bf ParsedSentence}.
% This style of modelling is similar
% to that employed by LaSIE%
% \footnote{Large-Scale Information Extraction.}%
% , the Sheffield NLP group's MUC-6 system (Gaizauskas, Humphreys, Wakao,
% Cunningham 1995). VIE (see section \ref{vie}) uses the TIPSTER model -- see
% below.

Finally, MULTEXT has developed some general support infrastructure for
handling SGML and for parallelising tool pipelines.
A query language for accessing components of SGML
documents is defined 
and API in support of this language provided. For example
a program might specify parts of a document by the pattern
{\tt DOC/*/s}
which refers to all \verb|<s>| objects under \verb|<DOC>| tags -- all
SGML-marked sentences in the document.
Additionally SGML-aware versions of various UNIX utilities are in
development. Parallel execution may be supported at the level of single tools
via a program that distributes pipelined operations over a set of networked
machines.

\subsubsection*{\label{tipster}TIPSTER II}

The TIPSTER programme in the US, currently in its second phase, has also
produced a data-driven architecture for NLP systems.
Like MULTEXT, TIPSTER
addresses specific forms of language processing, in this case information
extraction and document detection (or information retrieval -- IR). As will
become clear below, however, TIPSTER's approach is not restricted to
particular NL tasks.

Whereas in MULTEXT all information about a text is encoded in SGML, 
which is added by the tools, in
TIPSTER a text remains unchanged while information is stored in a separate
database in the form of {\em annotations}. Annotations associate portions of
documents (identified by sets of start/end byte offsets or {\em spans})
%
% \footnote{LaSIE also uses byte offsets, preserving tokenisation information
% as a pair of offsets into the original source and referring to token
% identifiers throughout the processing stages. Reconstruction of the text
% with additional markup -- e.g. for the MUC named entity or coreference tasks
% -- is then a  relatively simple matter given markup specifiers based on token
% identifiers.}%
% )
with
analysis information ({\em attributes}), e.g.: POS tags;  textual unit type;
template element. In this way the information built up about a text by NLP
(or IR) modules is kept separate from the texts themselves. In place of an
SGML DTD an {\em annotation type declaration} defines the information
present in annotation sets, for example a set of values for MUC-style
organisation template elements. Figure \ref{annotations_eg} shows an example
from (Grishman \cite{Gri96}). SGML I/O is catered for by API calls
to import and export SGML-encoded text.
%\begin{quote}
%The first example shows a single sentence and the result of three
%annotation procedures: tokenization with part-of-speech assignment,
%name recognition, and sentence boundary recognition.  Each token has
%a single attribute, its part of speech (pos), using the tag set from
%the Univ. of Pennsylvania Tree Bank;  each name also has a
%single attribute, indicating the type of name:  person, company, etc.
%\end{quote}
%
\small
\begin{figure}[!htbp]
\small
\begin{center}   
\begin{tabular}{|l|l|r|r|l|} \hline
\multicolumn{5}{|c|}{\em Text} \\
\multicolumn{5}{|c|}{{\tt Sarah savored the soup.}} \\
\multicolumn{5}{|c|}{{\tt 0...|5...|10..|15..|20}} \\ \hline \hline
\multicolumn{5}{|c|}{\em Annotations} \\
Id & Type  & \multicolumn{2}{c|}{Span} & Attributes \\
   &       & Start & End &                          \\ \hline
 1 & token & 0  & 5  & pos=NP \\
 2 & token & 6  & 13 & pos=VBD \\
 3 & token & 14 & 17 & pos=DT \\
 4 & token & 18 & 22 & pos=NN \\
 5 & token & 22 & 23 &  \\ \hline
 6 & name  & 0  & 5  & name\_type=person \\ \hline
 7 & sentence & 0 & 23 & \\ \hline
\end{tabular}
\end{center}
\caption{TIPSTER annotations example \label{annotations_eg}}
\end{figure}
\normalsize

The definition of annotations in TIPSTER forms part of an object-oriented
model that deals with inter-textual information as well as single texts.
Documents are grouped into {\em collections}, each with a database storing
annotations and document attributes such as identifiers, headlines etc.
Collections are the first-class entities in the architecture. The
model also describes elements of IE and IR systems relating to their use,
with classes representing queries and information needs.

\subsubsection*{Comparison of MULTEXT and TIPSTER}

Both projects propose architectures appropriate for LE, but there are a
number of significant differences. We discuss seven here, then note the
possibility of complimentary interoperation of the two.

\begin{enumerate}
\item
MULTEXT adds new information to documents
by augmenting an SGML stream; TIPSTER stores information remotely in a
dedicated database. This has several implications. Firstly, TIPSTER can
support documents on read-only media (e.g. CD-ROMs, which may be used for
bulk storage by organisations with large archiving needs, even though access
will then be slower than from hard disk; but note that a recent revision to
the specification allows for writeable documents). Secondly, TIPSTER
avoids the difficulties referred to earlier of representing graph-structured
information in SGML. From the point of view of efficiency, the original
MULTEXT model of interposing SGML between all modules implies a generation
and parsing overhead in each module. Later versions have replaced this model
with a pre-parsed representation of SGML to reduce this overhead. This 
representation will presumably be stored in intermediate files, which
implies an overhead from the I/O involved in continually reading and writing
all the data associated with a document to file. There would seem no reason
why these files should not be replaced by a database implementation,
however, with potential performance benefits from the ability to do I/O on
subsets of information about documents (and from the high level of
optimisation present in modern database technology).
\item
A related issue is storage overhead. TIPSTER is minimal in this respect, as
there is no inherent need to duplicate the source text (which also means
that it works naturally with read-only media like CD-ROMs). MULTEXT
potentially has to duplicate the source text at each intermediary stage,
although this might be ameliorated by shifting to a database implementation.
\item
TIPSTER's data architecture is application-neutral -- the objects in the model
are generic to all information that is associated with definite ranges
of text. (The more concrete aspects of the architecture to do with IE and
IR model the objects involved in user interaction with such systems.)
MULTEXT's model is tool-specific in that the classes of object that the
model describes are those processed by the tools envisaged
(although the underlying
representation language, SGML, is information-neutral).
\item
There is no easy way in an SGML-based system to differentiate sets of
results (i.e. sets of markup) by e.g. the program or user that originated
them. In general, storing information about the information present in an
SGML system (or {\em meta-information}) is messy. This is a problem for
MULTEXT but not for TIPSTER. A related point is that TIPSTER can easily
support multi-level access control via a database's protection mechanisms --
this is again not straightforward in SGML.
\item
Distributed control is easy to implement in a database-centred system like
TIPSTER -- the DB can act as a blackboard, and implementations can take
advantage of well-understood access control (locking) technology. How to do
distributed control in MULTEXT is not obvious.
\item
TIPSTER provides no tools or databases, but many sites are already committed
to TIPSTER-compatibility, so the set of modules available in the framework
will grow over time.
MULTEXT is based
around a set of tools and reference corpora annotated accordingly.
\end{enumerate}
Interestingly, a TIPSTER system could function as a module in a MULTEXT
system, or vice-versa. A TIPSTER storage system could write data in SGML for
processing by MULTEXT tools, and convert the SGML results back into native
format. Also, the extensive work done on SGML processing in MULTEXT could
usefully fill a gap in the current TIPSTER model, in which SGML capability
is not fully specified (plans are currently being formed in the US
to address this
problem -- input from European experience would seem advisable).
Integration of the results of both projects would
seem to be the best of both worlds, and we hope to achieve this in GATE.
%\resetparskip

% $Log: related.tex,v $
% Revision 1.2  1996/07/09  16:55:06  hamish
% final submission
%
% Revision 1.1  1996/04/12  13:05:53  hamish
% as sent to Oflazer

% \input{references}

\bibliographystyle{plain}
\bibliography{../utils/gentex}

\end{document}